%% file: acl2021.tex
\pdfoutput=1
\documentclass[11pt]{article}

\usepackage[]{acl2021}

\usepackage{times}
\usepackage{latexsym}

\usepackage[T1]{fontenc}
\usepackage[utf8]{inputenc}

\usepackage{microtype}

\usepackage{makecell}

\usepackage{microtype}

\usepackage{graphicx}
\usepackage{paralist}
\usepackage{url}
\usepackage{xspace}
\usepackage{amsfonts,amsmath}
\usepackage{booktabs}
\usepackage{setspace}
\usepackage{multirow}
\usepackage{multicol}
\usepackage{subfigure}
\usepackage{hyperref}
\usepackage{amssymb}% http://ctan.org/pkg/amssymb
\usepackage{pifont}% http://ctan.org/pkg/pifont
\usepackage{color, colortbl}
\usepackage{bbm}
\usepackage{wrapfig}
\usepackage{lipsum}
\usepackage{enumitem}
\usepackage{paralist, tabularx}
\usepackage{balance}

\newcommand\blfootnote[1]{%
  \begingroup
  \renewcommand\thefootnote{}\footnote{#1}%
  \addtocounter{footnote}{-1}%
  \endgroup
}

\title{Dict-BERT: Enhancing Language Model Pre-training with Dictionary}
\author{Wenhao Yu$^{1*}$, Chenguang Zhu$^{2}$, Yuwei Fang$^{2}$, Donghan Yu$^{3*}$, \\ \bf Shuohang Wang$^{2}$, Yichong Xu$^{2}$, Michael Zeng$^{2}$, Meng Jiang$^{1}$ \\
$^1$University of Notre Dame, Notre Dame, IN \\ $^2$Microsoft Cognitive Services Research, Redmond, WA \\ $^3$Carnegie Mellon University, Pittsburgh, PA  \\
{$^1$\tt \{wyu1, mjiang2\}@nd.edu}; 
{$^2$\tt chezhu@microsoft.com}
}

\begin{document}
\maketitle

\blfootnote{* This work was done when Wenhao Yu and Donghan Yu interned at Microsoft Cognitive Services Research group.}

\begin{abstract}
\input{0-abstract}

\end{abstract}

\section{Introduction}
\label{sec:introduction}
\input{1-introduction}

\section{Related Work}
\label{sec:related}
\input{2-relatedwork}

\section{Proposed Method}
\label{sec:method}
\input{3-method}

\section{Experiments}
\label{sec:Experiments}
\input{4-experiments}

\section{Conclusions}
\label{sec:conclusions}
\input{5-conclusion}

\section*{Acknowledgements}
Wenhao Yu and Meng Jiang are supported in part by the National Science Foundation IIS-1849816, CCF-1901059, IIS-2119531 and IIS-2142827.

\balance
\bibliography{reference}
\bibliographystyle{acl_natbib}

\clearpage
\appendix
\section{Appendix}
\input{appendix}

\end{document}

%% file: 0-abstract.tex
Pre-trained language models (PLMs) aim to learn universal language representations by conducting self-supervised training tasks on large-scale corpus. 
Since PLMs capture word semantics in different contexts, the quality of word representations highly depends on word frequency, which usually follows a heavy-tailed distribution in the pre-training corpus.
Thus, the embeddings of rare words on the tail are usually poorly optimized. 
In this work, we focus on enhancing language model pre-training by leveraging definitions of the rare words in dictionary.
To incorporate a rare word definition as a part of input,
we fetch it from the dictionary and append it to the end of the input text sequence.
In addition to training with the masked language modeling objective, we propose two novel self-supervised pre-training tasks on word-level and sentence-level alignment between the input text and rare word definition
% definition \shuo{text-definition matching still looks confusing. Not clear what's text.}matching 
to enhance language representations. % with dictionary.
%We propose a contrastive objective and a discriminative objective for learning knowledge from a dictionary and improving contextualized word representations, 
%and develop a joint pre-training framework that unifies the two self-supervised learning tasks with masked language model.
% with masked language models so that they can mutually enhance each other.
% and jointly learn the two self-supervised learning tasks for incrementally pre-training PLMs.
%In order to avoid the appended rare word definitions diverting the sentence from its original meaning,
% During fine-tuning, we employ a knowledge attention mechanism that makes the definitions only visible to the corresponding words in the input sequence. %\shuo{Is finetuning new? Maybe remove this sentence if not novel.}
% Our proposed methods can directly work with any existing PLM architecture.
We evaluate the proposed model named Dict-BERT on the GLUE benchmark and eight specialized domain datasets.
% Extensive experiments demonstrate that Dict-BERT can significantly boost model performance on various downstream tasks with better understanding of rare words.
Extensive experiments show that Dict-BERT significantly improves the understanding of rare words and boosts model performance on various NLP downstream tasks.

%% file: 1-introduction.tex
Recently pre-trained language models (PLMs) such as BERT~\citep{devlin2019bert} and RoBERTa~\citep{liu2019roberta} have revolutionized the field of natural language processing (NLP), yielding remarkable performance on various downstream tasks~\citep{qiu2020pre}.
% For example, BERT~\citep{devlin2019bert} and its novel variants such as RoBERTa~\citep{liu2019roberta} and XLNet~\citep{yang2019xlnet} capture syntactical and semantic knowledge mainly from the pre-training task of masked language modeling (MLM). 
However, these PLMs suffer from lacking knowledge when completing real-world tasks. To address this issue, some methods have incorporated the knowledge to enrich language representations, ranging from linguistic~\citep{wang2021k}, commonsense~\citep{guan2020knowledge,liu2020k}, factual~\citep{wang2021kepler}, to domain knowledge~\citep{liu2020k,yu2020jaket}.

Nevertheless, 
%the robustness of these knowledge-enhanced language models is still weak with regard to rare words~\citep{schick2020rare} and unseen words~\citep{cui2021knowledge} in downstream tasks.
rare words~\citep{schick2020rare} and unseen words~\citep{cui2021knowledge} are still blind spots of pre-trained language models when they are fine-tuned on downstream tasks.
For instance, in a dialogue system, users often talk to chatbots about recent hot topics, e.g., ``Covid-19'', which may not appear in the pre-training corpus~\citep{cui2021knowledge}.
Since PLMs capture word semantics in different contexts to address the issue of polysemous and the context-dependent nature of words, consequently they usually perform poorly when a user mentions such novel words~\citep{wu2021taking,ruzzetti2021lacking}.
% Since PLMs capture word semantics in different contexts to address the issue of polysemous and the context-dependent nature of words.
As indicated by \citet{wu2021taking}, the quality of word representations highly depends on the word frequency in the pre-training corpus, which typically follows a heavy-tail distribution. Thus, a large proportion of words appear very few times and the embeddings of these rare words are poorly optimized~\citep{gong2018frage,schick2020rare}. Such embeddings usually carry inadequate semantic meaning, which complicate the understanding of input text, and even hurt the pre-training of the entire model.

In this work, we focus on enhancing language model pre-training by leveraging rare word definitions in English dictionaries (e.g., Wiktionary). Definitions in dictionaries are intended to describe the meaning of a word to a human reader.
We append the definitions of rare words to the end of the input text and encode the whole sequence with Transformer encoder.
The pre-training tasks are mainly based on the alignment between input text and the appended word definitions, some of which are randomly sampled polluted words and don't explain the input.
%When encountering a rare word in the input text, we fetch its definition from dictionary and append it to the end of the input text.
%In addition of training with the masked language modeling (MLM) objective, we propose two novel pre-training tasks on word and sentence level text-definition alignment to enhance language modeling representation with dictionary.
We propose two types of pre-training objectives:
1) a word-level contrastive objective aims to maximize the mutual information between Transformer representations of a rare word appeared in the input text and its dictionary definition.
2) a sentence-level discriminative objective aims at learning to differentiate between correct and polluted word definitions.
% In order to make better interactions between the input text sequence and rare word definitions, we propose two novel self-supervised training tasks for learning knowledge from dictionary and improving contextualized word representations: (1) the contrastive objective aims to maximize the mutual information between input text sequence and dictionary definitions, and (2) the discriminative objective aims at learning to differentiate between correct and polluted word definitions.
% We develop a joint pre-training framework that unifies our proposed two self-supervised learning tasks with the MLM objective.
During downstream fine-tuning, in order to avoid the appended rare word definitions diverting the sentence from its original meaning, we employ a knowledge attention mechanism that makes word definitions only visible to the corresponding words in the input text sequence.
We name our method Dict-BERT. 
Notably, Dict-BERT is general and model-agnostic, in the sense that any pre-trained language model (e.g., BERT, RoBERTa) suffices and can be used.

% \yuwei{Shall we claim significantly improve the understanding of rare words here? Reviewers might look at the table 4, but there are no difference between the last three rows and the improvement is not that significant.}
% \yuwei{Shall we emphasize on the rare word? I think it is more about the implementation details and also our method can improve the general understanding ability. We can explain the rare word representation has been improved from Table 4 as one aspect.}

%We evaluate our proposed method on the language understanding benchmark GLUE~\citep{wang2019glue} and eight specialized domain benchmark datasets~\citep{gururangan2020don}.
%Extensive experimental results demonstrate that our method can significantly improve the understanding of rare words and boost model performance on various downstream tasks. For example, compared with the vanilla BERT, our method can improve accuracy by +1.15\% on the GLUE benchmark.

Overall, our main contributions in this work can be summarized as follows:
\begin{enumerate}[wide=5pt, itemsep=-0.5ex, topsep=-2pt,]
    \item We are the first work to enhance language model pre-training with rare word definitions from dictionaries (e.g., Wiktionary).
    \item We propose two novel pre-training tasks on word-level and sentence-level alignment between input text sequence and rare word definitions to enhance language modeling with dictionary. 
    \item We evaluate Dict-BERT on the GLUE~\cite{wang2019glue} benchmark, in which our model pre-trained from scratch can improve accuracy by +1.15\% on average over the vanilla BERT.
    \item We follow the domain adaptive pre-training (DAPT) setting~\citep{gururangan2020don}, where language models are continuously pre-trained with in-domain data. We evaluate Dict-BERT on eight specialized domain datasets. Our method can improve F1 score by +0.5\%/+0.7\% on average over the BERT-DAPT/RoBERTa-DAPT settings.
\end{enumerate}
%(1) We are the first work to integrate word definitions in dictionary into language model pre-training. (2) We propose two novel pre-training tasks on word and sentence level text-definition alignment to enhance language modeling representation with dictionary.
% and improving contextualized word representations. 
%(3) Our proposed Dict-BERT can outputperform BERT on the GLUE benchmark and BERT/RoBERTa under domain adaptive pre-training (DAPT) setting.

%% file: 2-relatedwork.tex
\paragraph{Rare word representation in language models.}
The quality of word representations highly depends on word frequency creating a heavy-tail distribution \cite{wu2021taking}.
Recent works have shown rare words that are not frequently covered in the corpus can hinder the understanding of specific yet important sentences~\citep{noraset2017definition,bosc2018auto,schick2020rare,ruzzetti2021lacking}. Due to the poor quality of rare word representations, the pre-training model built on top of it suffers from noisy input semantic signals which lead to inefficient training. 
\citet{gao2019representation} provided a theoretical understanding of the rare word problem, which illustrates that the problem lies in the sparse stochastic optimization of neural networks. 
\citet{schick2020rare} adapted attentive mimicking to explicitly learn rare word embeddings to language models. 
% Specifally, it introduces one-token approximation, a procedure that uses attentive mimicking even when the underlying language model uses subword-based tokenization.
% \citet{xu2020fusing} appends dictionary definitions to the input question and choice to boost language model's performance on commonsense question answering.
\citet{wu2021taking} proposed to maintain a note dictionary and saves a rare word’s contextual information as notes. When the same rare word occurs again during language model pre-training, the note information saved beforehand can be employed to enhance the semantics of the current sentence. 
Different from aforementioned works that keep a fixed vocabulary of rare words during pre-training and fine-tuning, our method can dynamically adjust the vocabulary of rare words, obtain and represent their definitions in a dictionary in a plug-and-play manner.
% By doing so, TNF provides better data utilization since cross-sentence information is employed to cover the inadequate semantics caused by rare words in the sentences.

% \paragraph{Knowledge-enhanced language model pre-training.}
% % Memory-augmented BERT. Another line of work close to ours uses memory-augmented neural
% % networks in language-related tasks. Fevry et al. (2020) and Guu et al. (2020) define the memory ´
% % buffer as an external knowledge base of entities for better open domain question answering tasks.
% % Khandelwal et al. (2019) constructs the memory for every test context at inference, to hold extra
% % token candidates for better language modeling. Similar to other memory-augmented neural networks,
% % the memory buffer in these works is a model component that will be used during inference. Although
% % sharing general methodological concepts with these works, the goal and details of our method are
% % different from them. Especially, our note dictionary is only maintained in pre-training for efficient
% % data utilization. At fine-tuning, we ditch the note dictionary, hence adding no extra time or space
% % complexity to the backbone models.

\paragraph{Language model pre-training and knowledge-enhanced methods}
Recent years have seen substantial pre-trained language models (PLMs) such as BERT~\citep{devlin2019bert} and T5~\citep{raffel2020exploring} have achieved remarkable performance in various NLP downstream tasks.
However, these PLMs suffer from lacking domain-specific knowledge when completing many real-world tasks~\citep{yu2020survey}.
For example, BERT cannot give full play to its value when dealing with electronic medical record analysis tasks in the medical field~\citep{liu2020k}.
A lot of efforts have been made on investigating how to integrate knowledge into PLMs~\citep{yu2020jaket,liu2021kg,xiong2020pretrained,guan2020knowledge,zhou2021pre,yu2021kg,yu2022diversifying}. 
Overall, these approaches can be grouped into two categories:
The first one is to explicitly inject knowledge representation into PLMs, where the representations are pre-computed from external sources~\citep{zhang2019ernie,liu2021kg}.
% For example, KG-BART encoded the graph structure of KGs with knowledge graph embedding algorithms and then took the informative entity embeddings as auxiliary input~\citep{liu2021kg}.
However, it has been argued that the embedding vectors of input words and knowledge are obtained in separate ways, making their vector-space inconsistent~\citep{liu2020k}.
The second one is to implicitly model knowledge information into PLMs by performing knowledge-related tasks, such as concept order recovering~\citep{zhou2021pre}, entity category prediction~\citep{yu2020jaket}.
However, none of existing work has explored using dictionary to enhance language model pre-training.
% For example, JAKET jointly pre-trained both the KG representation and language representation by adding two self-supervised learning objectives (i.e., entity category prediction, relation type prediction) on KGs~\citep{yu2020jaket}.
% For example, CALM proposed a novel contrastive objective for packing more commonsense knowledge into the parameters for enhancing language understanding tasks~\citep{zhou2021pre}.

%% file: 3-method.tex
In this section, we introduce the details of our model Dict-BERT. We first describe the notations and how to incorporate rare word definitions as a part of input. Then we detail the two novel self-supervised pre-training objectives. Finally, we introduce the knowledge attention during fine-tuning.

\subsection{Notation and Problem Definition}

Given the input text sequence $X = [\text{CLS}, x_1, x_2, \cdots, x_L, \text{SEP}]$ with $L$ tokens, a language model $f_{LM}$ produces the contextual word representation $f_{LM}(X) = [h_{\text{CLS}}, h_1, {h}_2, \cdots, {h}_L, h_{\text{SEP}}]$. For a specific downstream task, a header function $f_{H}$ further uses $f_{LM}(X)$ and generates the prediction as $f_H({h}_{\text{CLS}})$ for sequence classification tasks.
% or $f_H([h_{\text{CLS}}, {h}_1, {h}_2, \cdots, {h}_L, h_{\text{SEP}}])$ for token classification.

\vspace{0.05in}
\textit{The goal of our work} is to learn better contextual word representation $f_{LM}(x)$ by leveraging definitions of the rare words in dictionaries (e.g., Wiktionary). 
% \yx{We can define a set (e.g., $S_{\text{rare}}$) for the set of rare words.}
Suppose $S = [s_1, \cdots, s_K]$ and $C = [c^{(1)}, \cdots, c^{(K)}]$ are the sets of rare words in the input text sequence $X$ and their definitions in the dictionary.
When a rare word $s_i$ appears in the input text sequence, we fetch its definition from the dictionary as $c^{(i)} = [c_1^{(i)}, \cdots, c_{N_i}^{(i)}]$ with $N_i$ tokens, and append it to the end of the input text sequence.
If a word has multiple definitions, we use the definition of their first etymology (i.e., the most commonly used meaning).
Therefore, an input sequence $X$ with appended definitions of $K$ rare words can be written as: 
$\mathrm{[X; C]} = [\text{CLS}, x_1, x_2, ..., x_L, \text{SEP}^{(1)}, c_1^{(1)}, c_2^{(1)}, ..., c_{N_1}^{(1)}; ...;$  $\text{SEP}^{(K)}, c_1^{(K)}, c_2^{(K)}, ..., c_{N_K}^{(K)}, \text{SEP}]$, and the corresponding contextual representation generated from the language model $f_{LM}$ as: $f_{LM}(\mathrm{X, C}) = [h_{\text{CLS}}, {h}_1, {h}_2, \cdots, {h}_L, h_{\text{SEP}}^{(1)}, {h}_1^{(1)}, \cdots, {h}_{N_1}^{(1)}; \cdots \cdots$ $;h_{\text{SEP}}^{(K)}, {h}_1^{(K)}, \cdots, {h}_{N_K}^{(K)}, h_{\text{SEP}}]$. For a specific downstream task, a header function $f_{H}$ still uses $f_{LM}(\mathrm{X, C})$ to generate the prediction as $f_H({h}_{\text{CLS}})$ for sequence classification tasks. 
% or $f_H([h_{\text{CLS}}, {h}_1, {h}_2, \cdots, {h}_L])$ for token classification.

\subsection{Choosing the Rare Words}
\label{sec:rare}

% \yx{Do we have ablation for the two methods in this part? Does it matter? I would suggest putting 500 and 10\% as hyperparameters and specify them in appendix. Otherwise people will question our choice of them.}

 There are different ways to choose the rare word set $S$ in a pre-training corpus. One way is to use a pre-defined absolute frequency value as the threshold. \citet{wu2021taking} used 500 as the threshold to divide frequent words and rare words, and maintained a fixed vocabulary of rare words during pre-training and fine-tuning. However, rare words can vary greatly in different corpora. 
For example, rare words in the medical domain are very different from those in general domain \citep{lee2020biobert}.
Besides, keeping a large threshold for a small downstream datasets makes the vocabulary of rare words too large.
For example, only 51 words in the RTE dataset have a frequency of more than 500.
% e.g., MNLI has 393k training data but RTE only has 2.5k training data, and 

Therefore, we propose to choose specialized rare words for each pre-training corpus and downstream tasks.
Specifically, 
we ranked all word frequency from smallest to largest, and add them to the list one by one until the word frequency of the added word reaches 10\% of the total word frequency. 
Compared with \citet{wu2021taking} which maintained a fixed vocabulary, our method can dynamically adjust the vocabulary of rare words, obtain and represent their definitions in dictionary in a plug-and-play manner.
%However, it is unrealistic to heuristically determine an appropriate threshold for each downstream task dataset since the size of different datasets varies greatly. In GLUE benchmark, for example, MNLI has 393k training data but RTE only has 2.5k training data. 
%Therefore, instead of using an absolute frequency value, the second way is to dynamically select rare words based on the word frequency distribution. 
To fetch the definition of rare words, we leveraged the largest online dictionary, i.e., Wiktionary, and collected a dump of Wiktionary which includes definitions of 999,614 concepts. 
%We noted that using above two rare word collection methods achieved very close performance in the BERT pre-training experiments. The main drawback of the first method still lies in choosing appropriate thresholds for each pre-training corpus and downstream task datasets. Therefore, we used the second method to collect rare words in our experiments. 

We noted that when choosing the rare words, we used a word tokenizer (i.e., NLTK) instead of using any subword tokenizer (e.g., WordPiece). This is mainly because quite a few rare subwords, either generated by BPE or in WordPiece, do not have specific understandable semantic meanings to humans, such as ``123@@'', ``elids'', ``al'', ``ch'', ``di''. For such subwords, their contexts can be very diverse due to their vague semantic meanings. As most rare words have their own concrete semantics, the subword meanings cannot act as effective auxiliary semantics to enhance the current input.

% \subsection{Preliminary: BERT Pre-training}

% BERT uses the Transformer model as its backbone neural network architecture and trains the model parameters with the masked language modeling (MLM) objective on large text corpora. 
% In the masked language modeling task, a random sample of the words in the input text sequence is selected.
% The selected positions will be either replaced by special token [MASK], replaced by randomly picked tokens or remain the same.
% The objective of masked language modeling is to predict words at the masked positions correctly given the masked sentences. 
% RoBERTa (robustly optimized BERT approach) is a retraining of BERT with improved training methodologies, 1000\% more data (i.e., 160 GB) and computation power (i.e., 1024 V100 GPUs).
% To improve the training procedure, RoBERTa introduces dynamic masking so that the masked token changes during the training epochs. 
% Larger batch-training sizes were also found to be more useful in the training procedure.

\begin{figure*}[t]
    \centering
    {\includegraphics[width=1.0\textwidth]{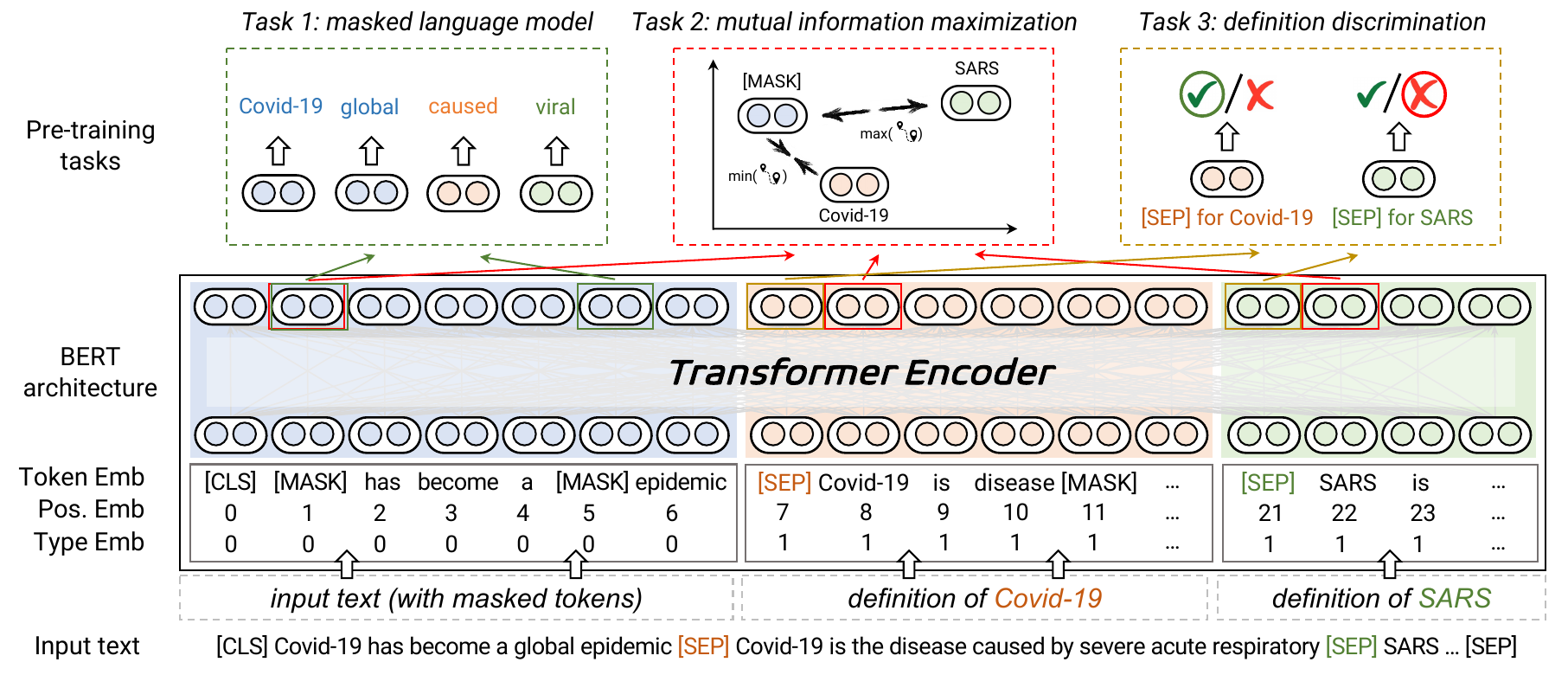}}
    \vspace{-0.25in}
    \caption{The overall architecture of Dict-BERT. The definitions of rare words are appended to the end of input text. In additional to training with masked language modeling, Dict-BERT performs two novel self-supervised learning tasks: word-level mutual information maximization ($\S$\ref{sec:mmi}) and sentence-level definition discrimination ($\S$\ref{sec:dhp}). }
    % ``SARS'' is a negatively sampled rare word.}
    \vspace{-0.1in}
    \label{fig:framework}
\end{figure*}

\subsection{Dict-BERT: Language Model Pre-training with Dictionary}

Dict-BERT is based on the BERT architecture, which can be initialized either randomly or from a pre-trained checkpoint with the same structure. It is worth noting that we slightly modified the type embedding, in which the type embedding of the input text is set as $0$, and the type embedding of the dictionary definitions is set as $1$. In addition, we used the absolute positional embedding.

We represent each input text sequence and dictionary definitions pair as a tuple $(X, C)$.
The semantics of a word in the input text depends on the current context, while the semantics of a word in the dictionary is standardized by linguistic experts.
In order to better align the representations between them, we propose two novel pre-training tasks on word-level and sentence-level alignment between input text sequence and rare word definitions to enhance pre-trained language models with dictionary.

% \subsubsection{Maximum Mutual Information between Input Text and Word Definition}
\subsubsection{Word-level Mutual Information Maximization}
\label{sec:mmi}

Recently, there has been a revival of approaches inspired by the InfoMax principle~\citep{oord2018representation,tschannen2020mutual}: maximizing the mutual information (MI) between the input and its
representation. MI measures the amount of information obtained about a random variable by observing another random variable. As the input text sequence and rare word definitions are obtained from different sources, in order to better align their semantic representations, we proposed to maximize the MI between a rare word $x_i$ in the input sequence and its well-defined meaning in the dictionary $c^{(i)}$, with joint density $p(x_i, c^{(i)})$ and marginal densities $p(x_i)$ and $p(c^{(i)})$, is defined as the Kullback–Leibler (KL) divergence between the joint and the product of the marginals, 
\begin{align}
    I(x_i; c^{(i)}) & = D_{KL} \big{(}p(x_i, c^{(i)})||p(x_i)p(c^{(i)}) \big{)} 
    % \nonumber 
    % \\ & = \mathbbm{E}_{p(x_i, c^{(i)})}[\log \frac{p(x_i, c^{(i)})}{p(x_i)p(c^{(i)})}]. 
\end{align}
% As both the input sequence $X$ and the dictionary definition $C$ describe the same word, we aim to maximize $I(x_i; c^{(i)})$.
The intuition of maximizing mutual information between a rare word appeared in the input text sequence and its definitions in the dictionary is to encode the underlying shared information and align the semantic representation between the contextual meaning and well-defined meaning of a word.
Nevertheless, estimating MI in high-dimensional spaces is a notoriously difficult task, and in practice one often maximizes a tractable lower bound on this quantity~\citep{poole2019variational}. Intuitively, if a classifier can accurately distinguish between samples drawn from the joint $p(x_i, c^{(i)})$ and those drawn from the product of marginals $p(x_i)p(c^{(i)})$, then $x_i$ and $c^{(i)}$ have a high mutual information.

In order to approximate the mutual information, we adopted InfoNCE~\citep{oord2018representation}, which is one of the most commonly used estimators in the representation learning literature, defined as
\begin{align}
    I(x_i;c^{(i)}) & \geq \mathbbm{E}[ \sum_{i=1}^{K} \log \frac{e^{f_{\text{MI}}(h_{i}, h^{(i)})}}{\sum_{j=1}^{K} \mathbbm{1}_{[j \neq i]} e^{f_{\text{MI}}(h_{i}, h^{(j)})}}] \nonumber \\ & \triangleq I_{NCE}(x_i; c^{(i)}),
    \label{eq:mmi-ob}
\end{align}
where the expectation is over $K$ independent samples $\{(h_{i}, h^{(i)})\}^{K}_{i=1}$ from the joint distribution $p(x_i, c^{(i)})$ \citep{poole2019variational}. Intuitively, the critic function $f_{\text{MI}}(\cdot)$ measures the similarity (e.g., inner product) between two word representations.
% tries to predict the contextual rare word representation $h_{i}$ using its corresponding definition embedding: $h^{(1)}, \cdots ,$ or $h^{(K)}$. 
The model should assign high values to the positive pair $(h_{i}, h^{(i)})$, and low values to all negative pairs. We compute InfoNCE using Monte Carlo estimation by averaging over multiple batches of samples~\citep{chen2020simple}.
By maximizing the mutual information between the encoded representations, we extract the underlying latent variables that the rare words in the input text sequence and their dictionary definitions have in common.

\subsubsection{Sentence-level Definition Discrimination}
% \shuo{discriminative loss vs contrastive loss}}
\label{sec:dhp}

Instead of locally aligning the semantic representation, learning to differentiate between correct and polluted word definitions helps the language model capture global information of input text and dictionary definitions.
% \cz{together with previous sections, please define clearly: input text (whole) h?, specific rare word x, x's embedding, that word's definition text, that word's definition embedding}
We denote the set of definitions of rare words in the input text as ${C}$.
We then create a set of ``polluted'' word that are randomly sampled from the entire vocabulary together with its definition.
The number of sampled ``polluted'' words is equal to the number of rare words appeared in the input text sequence.
\begin{equation}
    \mathcal{L}_{\text{DD}} = - \mathbbm{E} \sum_{i=1}^{K} \log p(y|f_{\text{MLP}}(h_{\text{SEP}}^{(i)}).
\end{equation}

\subsubsection{Overall objective.} 
Now we present the overall training objective of Dict-BERT. To avoid catastrophic forgetting~\citep{mccloskey1989catastrophic} of general language understanding ability, we train the masked language modeling together with word-level mutual information maximization (MIM) and definition discrimination (DD) tasks. We denote $\mathcal{L}_{\text{MIM}}$ as the loss function of the MIM task which is the opposite of expectation in Equation \ref{eq:mmi-ob}. Hence, the overall learning objective is formulated as:
\begin{align}
    \mathcal{L} = \mathcal{L}_{\text{MLM}}+ \lambda_1 \mathcal{L}_{\text{MIM}} + \lambda_2  \mathcal{L}_{\text{DD}}
\end{align}
where $\lambda_1$, $\lambda_2$ are introduced as hyperparameters to control the importance of each task.

\begin{figure}
  \begin{center}
    \includegraphics[width=0.4\textwidth]{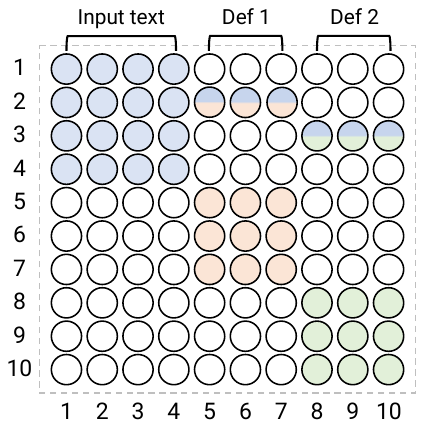}
  \end{center}
  \vspace{-0.15in}
  \caption{An illustration of knowledge-visible attention matrix. ``Def 1'' is the dictionary definition of the second word in the input text, and ``Def 2'' is the definition of the third word in the input text. Colored circle means token $i$ can attend information from token $j$, while white circle means no attention from token $i$ to token $j$. }
  \vspace{-0.15in}
  \label{fig:kvattn}
\end{figure}

\subsection{Dict-BERT: Fine-tuning with Knowledge-visible Attention}
\label{sec:fine-tune}
Most existing work uses the final hidden state of the first token (i.e., the [CLS] token) as the sequence representation~\citep{devlin2019bert,liu2019roberta,yang2019xlnet}. 
For a sequence classification task, a multi-layer perception network function $f_{H}$ takes the output of $f_{LM}$ as input and generates the prediction as $f_H({h}_{\text{CLS}})$.
Notably, when fine-tuning a language model on downstream tasks, there could be many rare/unseen words in the dataset. So, in the fine-tuning stage, when encountering a rare word in the input text, we append its definition to the end of input text, just like what we did in pre-training.
 
However, the appended dictionary definitions may change the meaning of the original sentence since the [CLS] token attend information from both input text and dictionary description.
As pointed in \citet{liu2020k} and \citet{xu2021does}, too much knowledge incorporation may divert the sentence from its original meaning by introducing a lot of noise. This is more likely to happen if there are multiple rare words in the input text. To address this issue, we adopt the visibility matrix \citep{liu2020k} to limit the impact of definitions on the original text. In BERT, an attention mask matrix is added with the self-attention weights before $\mathrm{softmax}$. 
If token $j$ is not supposed to be visible to token $i$, we add a -$\infty$ value in the attention matrix ($i$, $j$). 

As shown in Figure \ref{fig:kvattn}, we modify the attention mask matrix such that a token $i$ can attend to another token $j$ only if: (1) both tokens belong to the input text sequence, or (2) both tokens belong to the definition of the same rare word, or (3) $i$ is a rare word in the input text sequence and $j$ is from its definition in the dictionary.

%% file: 4-experiments.tex
\subsection{Tasks and Datasets}
To show the wide adaptability of our Dict-BERT, we conducted experiments on 16 NLP benchmark datasets. We use BERT~\cite{devlin2019bert} and RoBERTa~\cite{liu2019roberta} as the backbone pre-trained language methods. First, we followed \citet{liu2019roberta} and \citet{wu2021taking} to use 8 natural language understanding tasks in GLUE, including CoLA, RTE, MRPC, STS, SST, QNLI, QQP, and MNLI. 
Second, we followed \citet{gururangan2020don} to use 8 specialized domain tasks, including Chemprot, RCT-20k, ACL-ARC, SciERC, HyperPartisan, AGNews, Helpfulness, IMDB.

\begin{table*}[t]
\begin{center}
\caption{{Performance of different models on GLUE tasks. Each configuration is run five times with different random seeds, and the average of these five results on the validation set is reported in the table.
\textit{We note that} our code is implemented on Huggingface Transformer~\citep{wolf2020transformers}. The performance of our implemented BERT is consistent with the official performance, but it is slightly lower than the performance reported by \citet{wu2021taking}. We reported the relative improvement ($\Delta$) of BERT-TNF and Dict-BERT compared with the original BERT.}}
\vspace{-0.1in}
\setlength{\tabcolsep}{1.8mm}{\scalebox{0.89}{\begin{tabular}{l||cc|cccccccc|cc}
\toprule
{\multirow{2}*{Methods}} & \multicolumn{2}{c|}{{Dict in}} & MNLI & QNLI & QQP & SST & CoLA & MRPC & RTE & STS-B & {\multirow{2}*{Avg}} & {\multirow{2}*{$\Delta$}} \\
\cmidrule{2-11}
& PT & FT & Acc. & Acc. & Acc. & Acc. & Matthews & Acc. & Acc. & Pearson & &  \\
% \midrule
% BERT $\S$ & $\times$ & $\times$ & 85.00 & 91.50 & 91.20 & 93.30 & 58.30 & 88.30 & 69.00 & 88.50 & 83.10 & - \\
% BERT-TNF $\S$ & $\surd$ & $\surd$ & 85.00 & 91.00 & 91.20 & 93.20 & 59.50 & 89.30 & 73.20 & 88.50 & 83.90 & +0.80 \\
\midrule
{BERT (Wu's)} & {$\times$} & {$\times$} & {85.00} & {91.50} & {91.20} & {93.30} & {58.30} & {88.30} & {69.00} & {88.50} & {83.10} & {-} \\
{BERT-TNF} & {$\surd$} & {$\surd$} & {85.00} & {91.00} & {91.20} & {93.20} & {59.50} & {89.30} & {73.20} & {88.50} & {83.90} & {+0.80} \\
\midrule
BERT (ours) & $\times$ & $\times$ & 84.12 & 90.69 & 90.75 & 92.52 & 58.89 & 86.17 & 68.67 & 89.39 & 82.65 & - \\
Dict-BERT-F & $\times$ & $\surd$ & 84.19 & 90.94 & 90.68 & 92.59 & 59.16 & 85.75 & 68.10 & 88.72 & 82.51 & -0.14 \\
\midrule
Dict-BERT-P & $\surd$ & $\times$ & 84.33 & 91.02 & 90.69 & 92.62 & 60.44 & 86.81 & \textbf{73.86} & \textbf{89.81} & 83.70 & +1.05 \\
~$\vdash$ w/o MIM & $\surd$ & $\times$ & 84.24 & 90.79 & 90.24 & 92.22 & 60.14 & 87.03 & 73.79 & 89.67 & 83.52 & +0.87 \\  
~$\vdash$ w/o DD & $\surd$ & $\times$ & 84.18 & 90.54 & 90.30 & 92.39 & 61.49 & 86.49 & 71.89 & 89.60 & 83.36 & +0.71 \\  
\midrule
Dict-BERT-PF & $\surd$ & $\surd$ & \textbf{84.34} & \textbf{91.20} & \textbf{90.81} & \textbf{92.65} & \textbf{61.68} & \textbf{87.21} & 72.89 & 89.68 & \textbf{83.80} & \textbf{+1.15} \\
~$\vdash$ w/o MIM & $\surd$ & $\surd$ & 84.22 & 90.67 & 90.66 & 92.53 & 61.58 & 87.20 & 71.58 & 89.37 & 83.47 & +0.82 \\  
~$\vdash$ w/o DD & $\surd$ & $\surd$ & 84.16 & 90.21 & 90.78 & 92.39 & 61.14 & 87.19 & 71.84 & 89.24 & 83.37 & +0.72 \\  
\bottomrule
\end{tabular}}}
\label{tab:dictbert-glue}
\end{center}
\end{table*}

\begin{table*}[t]
\begin{center}
\caption{Performance of different models on eight specialized domain datasets under the domain adaptive pre-training (DAPT) setting. Each configuration is run five times with different random seeds, and the average of these five results on the test set is calculated as the final performance. }
% We also compared with other domain-specific pre-trained language models (e.g., BioBERT), which can be found at Table \ref{tab:domain-specialized} in Appendix.}
\vspace{-0.1in}
\setlength{\tabcolsep}{1.7mm}{\scalebox{0.89}{\begin{tabular}{l||cccccccc|c}
\toprule
{\multirow{2}*{Methods}} & ChemProt & RCT & ACL-ARC & SciERC & HP & AGNews & Helpful & IMDB & {\multirow{2}*{Avg}} \\
\cmidrule{2-9}
& Mi-F1 & Mi-F1 & Ma-F1 & Ma-F1 & Ma-F1 & Ma-F1 & Ma-F1 & Ma-F1 \\
% \midrule
% State-of-the-art & 84.60 & 92.90 & 71.00 & 81.80 & 94.80 & 95.50 & 65.10 & 96.20 & - \\
\midrule
BERT  & 81.16 & 86.91 & 64.20 & 80.40 & 91.17 & 94.48 & 69.39 & 93.67 & 82.67 \\  
% BERT-DAPT $\dag$ & 82.73 & 87.26 & 69.53 & 81.54 & - & - & - & - & - \\
BERT-DAPT  & 83.10 & 86.85 & 71.45 & 81.62 & 93.52 & \textbf{94.58} & 70.73 & 94.78 & 84.57 \\
Dict-BERT-DAPT & 83.49 & \textbf{87.46} & \textbf{74.18} & \textbf{83.01} & 94.70 & \textbf{94.58} & 70.04 & \textbf{94.80} & \textbf{85.25} \\
~$\vdash$ w/o MIM & 83.33 & 87.38 & 72.26 & 82.70 & \textbf{94.72} & \textbf{94.58} & 70.33 & 94.73 & 85.06 \\  
~$\vdash$ w/o DD & \textbf{84.09} & 87.23 & 72.78 & 82.54 & 94.69 & 94.57 & \textbf{70.43} & 94.70 & 85.01 \\  
\midrule
% RoBERTa $\S$ & 81.90 & 87.20 & 63.00 & 77.30 & 86.60 & 93.90 & 65.10 & 95.00 & 81.25 \\
RoBERTa & 82.03 & 87.14 & 66.20 & 79.55 & 90.15 & 94.43 & 68.35 & 95.16 & 83.15 \\
% RoBERTa-DAPT $\S$ & 84.20 & 87.60 & 75.40 & 80.80 & 88.20 & 93.90 & 66.50 & 95.40 & 83.95 \\
RoBERTa-DAPT & 84.02 & \textbf{87.62} & 73.56 & 81.85 & 90.22 & 94.51 & 69.06 & 95.18 & 84.51 \\
Dict-RoBERTa-DAPT & 84.41 & 87.42 & \textbf{75.33} & \textbf{82.53} & 92.51 & \textbf{94.80} & 70.57 & \textbf{95.51} & \textbf{85.32} \\
~$\vdash$ w/o MIM  & \textbf{84.49} & 87.51 & 74.83 & 81.58 & \textbf{93.27} & 94.75 & 70.67 & 95.40 & 85.31 \\  
~$\vdash$ w/o DD & 84.09 & 87.39 & 74.04 & 81.18 & 90.91 & 94.64 & \textbf{70.81} & \textbf{95.51} & 84.82 \\  
\bottomrule
\end{tabular}}}
\label{tab:dictbert-dsp}
\end{center}
\end{table*}

\subsection{Rare Word Collection}
\label{sec:rare_word}
Here, we briefly introduce the statistic of rare words in BERT pre-training corpus: English Wikipedia and BookCorpus. By concatenating these two datasets, we obtained a corpus with roughly 16GB in size. The total number of unique words in the pre-training corpus is 504,812, of which 112,750 (22.33\%) words are defined as frequent words. In other words, the sum of the occurrences of these 112,750 words in the corpus accounts for 90\% of the occurrences of all words in the corpus. We look up definitions of the remaining 392,062 (77.67\%) words in the Wiktionary, of which 252,581 (64.42\%) can be found. The average length of definition is 11.51$_{\pm6.84}$ words.

\subsection{Pre-training Corpus and Tasks}

\vspace{0.02in}
\textbf{Experiments on the GLUE benchmark.} The language model is first pre-trained on the general domain corpus, and then fine-tuned on the training set of different GLUE tasks. Following BERT~\citep{devlin2019bert}, we used the English Wikipedia and BookCorpus as the pre-training corpus. We removed the next sentence prediction (NSP) as suggested in RoBERTa~\citep{liu2019roberta}, and kept masked language modeling (MLM) as the objective for pre-training a vanilla BERT.

\vspace{0.02in}
\noindent\textbf{Experiments on specialized domain datasets.} The language model is not only pre-trained on the general domain corpus, but also pre-trained on domain specific corpus before fine-tuned on domain specific tasks. We initialized our model with the checkpoint from pre-trained BERT/RoBERTa and continue to pre-train on domain-specific corpus \citep{gururangan2020don}. The four domains we focus on are biomedical science (BIOMED), computer science (S2ORC-CS), news text (REALNEWS), and e-commerce reviews (AMAZON).

\subsection{Baseline Methods}

\vspace{0.02in}
\textbf{Vanilla BERT/RoBERTa.} We use the off-the-shelf BERT-base~\citep{devlin2019bert} and RoBERTa-base~\citep{liu2019roberta} model and perform supervised fine-tuning for each downstream tasks.

\vspace{0.02in}
\noindent\textbf{BERT-DAPT/RoBERTa-DAPT}. It continues pre-training BERT/RoBERTa on a large unlabeled domain-specific corpus (e.g., BioMed, RealNews) by MLM objective~\citep{gururangan2020don}. 

\vspace{0.02in}
\noindent\textbf{BERT-TNF}. It takes notes for rare words on the fly during pre-training to help the model understand them when they occur next time. Specifically, it maintains a note dictionary and saves a rare word’s contextual information in it as notes when the rare word occurs in a sentence~\citep{wu2021taking}.

\subsection{Implementation Details}
We introduce our pre-training and fine-tuning details and hyperparameter choices in Appendix \ref{sec:pre-training-details} to
\ref{sec:fine-tuning-details}. We also listed several detail discussions about using Wiktioanry in Appendix \ref{sec:wiktionary}.

\subsection{Ablation Settings}

\textbf{Dict-BERT-F} means that we load the vanilla BERT checkpoint and fine-tune on the downstream tasks by using knowledge attention for dictionary.

\vspace{0.02in}
\noindent\textbf{Dict-BERT-P} means that we only leverage dictionary in the pre-training stage and fine-tune Dict-BERT on downstream tasks without dictionary. 

\vspace{0.02in}
\noindent\textbf{Dict-BERT-PF} indicates that we use dictionary in both pre-training and fine-tuning stages.

\vspace{0.02in}
Furthermore, Dict-BERT w/o MIM removes the word-level mutual information maximization task and Dict-BERT w/o DD removes the sentence-level definition discriminative task during pre-training.

\begin{figure*}[t]
	\centering
	\subfigure[Full attn. (FT) v.s. Knowledge attn. (KT)]
	{\includegraphics[width=0.45\textwidth]{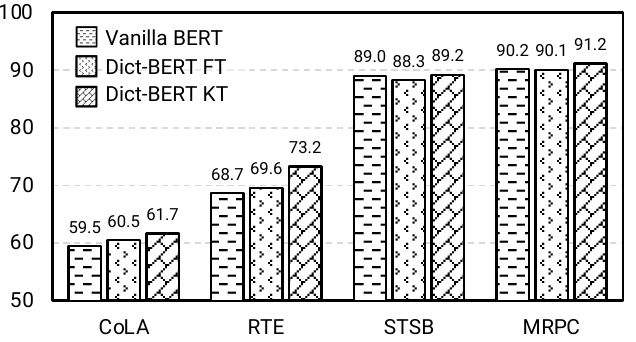}\label{fig:ktvsft}}
	\hspace{0.1in}
	\subfigure[Rare word ratios (5\% v.s. 10\% v.s. 15\%)]
	{\includegraphics[width=0.45\textwidth]{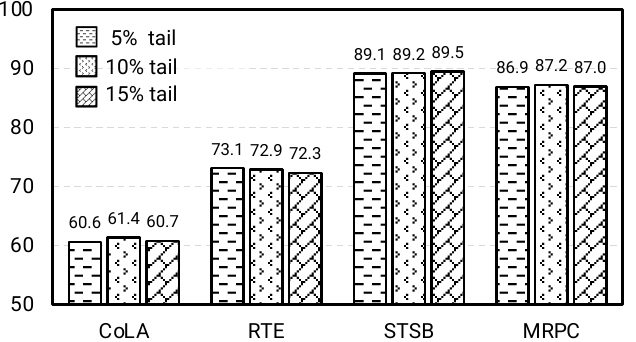}\label{fig:ratio}}
	\vspace{-0.15in}
	\caption{Model performance on CoLA, RTE, STSB and MRPC when (a) using two different attention mechanisms and (b) selecting different rare word ratios on the downstream task datasets during fine-tuning.}
\end{figure*}

\subsection{Experimental Results}

\vspace{0.02in}
\textbf{Dict-BERT-F v.s. BERT.} 
As shown in Table \ref{tab:dictbert-glue}, comparing the vanilla BERT with Dict-BERT-F, we observed that only using dictionary during fine-tuning could even hurt the model performance on the GLUE benchmark, especially on those small datasets (e.g., RTE, MRPC). This indicated the existing pre-trained language models cannot better understand the input sequence by using word definitions when not pre-trained with dictionary.
They might be even misled by the noisy explanations in the dictionary. 
Thus, it is important to incorporate  dictionary into language model pre-training so the dictionary definitions can be better utilized. 

\vspace{0.02in}
\noindent\textbf{Dict-BERT-PF v.s. BERT.} As shown in Table \ref{tab:dictbert-glue}, Dict-BERT-PF outperformed the vanilla BERT on the GLUE benchmark by improving +1.15\% accuracy on average. This indicated leveraging word definitions in dictionary can improve language model pre-training and boost performance on various NLP downstream tasks.
% The BERT performance from \cite{wu2021taking} is higher than our implemented BERT, however, they do not have open-source code for reproducing their experimental results. Though Dict-BERT-PF and BERT-TNF achieved very close performance on GLUE benchmark, i.e., 83.80\% and 83.90\%, our Dict-BERT-PF has achieved greater relative improvement on the GLUE benchmark than BERT-TNF, i.e., +1.15\% and +0.80\%. In addition, BERT-TNF keeps a fixed note dictionary so it cannot update any unseen words into the note dictionary during fine-tuning. On the contrary, Dict-BERT can dynamically adjust the vocabulary of rare words, obtain and represent their definitions in dictionary in a plug-and-play manner.
On RTE, Dict-BERT-P obtained the biggest performance improvement compared with the vanilla BERT. On another small-data sub-tasks CoLA, Dict-BERT-PF also outperformed the baseline with considerable margins. This indicated when Dict-BERT was fine-tuned on a small downstream dataset, the improvement was particularly significant. Besides, as shown in Table \ref{tab:dictbert-dsp}, Dict-BERT-DAPT outperformed BERT-DAPT on the specialized domain datasets by improving +0.68\% F1 on average. The same observation was obtained from the RoBERTa setting.
% comparison between Dict-RoBERTa-DAPT and RoBERTa-DAPT.

\vspace{0.02in}
\noindent\textbf{Dict-BERT-PF v.s. Dict-BERT-P.}
As shown in Table \ref{tab:dictbert-glue}, we compared model performance between using dictionary in fine-tuning (i.e., Dict-BERT-PF) and not using dictionary in fine-tuning (i.e., Dict-BERT-P). 
First, after pre-training the language model with dictionary, even without using dictionary in fine-tuning, the performance has been greatly improved. This indicated pre-training language model with dictionary generally improved the language representation and provided better initiation before fine-tuning the language model on the downstream tasks. 
Besides, we also observed the performance of Dict-BERT-PF performed slightly better than Dict-BERT-P. We hypothesized the reason behind can be the distribution discrepancy of the pre-training and fine-tuning data.

\vspace{0.02in}
\noindent\textbf{Ablation study.} As shown in Table \ref{tab:dictbert-glue} and Table \ref{tab:dictbert-dsp}, we conducted ablation study on both GLUE benchmark and specialized domain datasets. First, both MIM and DD helped learn knowledge from dictionary and improve language model pre-training. Specifically, DD demonstrated larger average improvement than MIM. The average improvements on GLUE benchmark brought by DD and MIM are +0.63\% and +0.52\%. Second, combining MIM and DD together achieved the highest performance on the GLUE benchmark, in which the average gain enlarges to +1.15\%. For specialized domain datasets, we had the same observations as above.

\vspace{0.02in}
\noindent\textbf{Knowledge attention v.s. Full attention.} As we mentioned in the Section \ref{sec:fine-tune}, too much knowledge incorporation may divert the sentence from its original meaning by introducing some noise. This is more likely to happen if there are multiple rare words appeared in the input text. Therefore, we compared the model performance between using knowledge attention and full attention. 
As shown in Figure \ref{fig:ktvsft}, we observed that using knowledge attention can consistently perform better than using full attention mechanism during the fine-tuning stage on CoLA, RTE, STSB and MRPC datasets. Besides, Dict-BERT with full attention even under-performed than the vanilla BERT without using any dictionary definition, which indicates the appended description in the dictionary may change the meaning of the original sentence. For example, STSB compares similarity between two sentence. Using full attention includes semantic meanings of definitions into the sentence representation, which might reduce the sentence similarity score and hurt the model performance.

\vspace{0.02in}
\noindent\textbf{Learning with different rare word ratios.} As we mentioned in Section \ref{sec:rare}, we select rare words for each downstream tasks by truncating the tail distribution of the word frequency. In order to verify the impact of using different tail proportions of rare words on the downstream tasks, we selected three different ratios (i.e., 5\%, 10\%, and 15\%) and experimented on CoLA, RTE, STSB and MRPC datasets. As shown in Figure \ref{fig:ratio}, on the CoLA and STSB datasets, the model achieves the best performance when using 10\% words at the tail as rare words. On the MRPC data, there is no significant difference of model performance in using different proportions of rare words. However, the performance on RTE data demonstrates a trend, that is, the more rare words selected, the worse the performance of the model. This is consistent with the conclusion of whether the dictionary is used in fine-tuning in Table \ref{tab:dictbert-glue}, i.e., the performance of not using dictionary is better than using dictionary on the RTE dataset. 
Thus, the selection of rare words with different tails has no obvious correlation with the performance of the model on downstream tasks.

\begin{table}[t]
\caption{Performance of different models on WNLaMPro test set, subdivided by word frequency.}
\vspace{-0.1in}
\centering
\scalebox{0.75}{\begin{tabular}{l||ccc|ccc}
\toprule
{\multirow{2}*{Methods}} & \multicolumn{3}{c|}{\textsc{Rare} (0, 10)} & \multicolumn{3}{c}{\textsc{Frequent} (100, +$\infty$)} \\
& MRR & P@3 & p@10 & MRR & P@3 & p@10 \\
\midrule
BERT (base) & 0.117 & 0.053 & 0.036 & 0.356 & 0.179 & 0.116 \\  
Dict-BERT & \textbf{0.145} & \textbf{0.068} & \textbf{0.041} & \textbf{0.359} & \textbf{0.181} & \textbf{0.117} \\  
~$\vdash$ w/o MIM & 0.144 & 0.067 & \textbf{0.041} & 0.357 & 0.180 & 0.115 \\
~$\vdash$ w/o DD & 0.141 & 0.065 & 0.040 & 0.355 & 0.179 & 0.116\\ \bottomrule
\end{tabular}}
\vspace{-0.1in}
\label{tab:wnlapro}
\end{table}

\vspace{0.02in}
\noindent\textbf{Unsupervised language model probing.} In order to assess the ability of language models to understand words as a function of their frequency, we used WordNet Language Model Probing (WNLaMPro) dataset~\citep{schick2020rare} to test how well a language model understands a given word: we can ask it for properties of that word using natural language. For example, a language model that understands the concept of ``guilt'', should be able to correctly complete the sentence ``Guilt is the opposite of \_\_\_'' with the word ``innocence''.
WNLaMPro contains four different kinds of relations: antonym, hypernym, cohyponym+, and corruption. Based on the word frequency in English Wikipedia, WNLaMPro defines three subsets based on keyword counts: \textsc{Rare} $(0, 10)$, \textsc{Medium} $(10, 100)$, and \textsc{Frequent} $(100, +\infty)$. As shown in Table \ref{tab:wnlapro}, Dict-BERT can greatly improve the word representation compared with the vanilla BERT without using a dictionary during pre-training. Based on the word frequency, we observe Dict-BERT can significantly help learn rare word representations. Compared to the vanilla BERT, Dict-BERT improves MRR and P@3 by relatively +23.93\% and +28.30\%, respectively. In addition, Dict-BERT is also able to learn better frequent word representations. Although we did not directly take frequent word definitions as part of the input, Dict-BERT spends less memory on rare words, because it is easier to predict rare words than the vanilla BERT, so the saved memory power could be used to memorize the facts involving popular words and interactions between popular words.

%% file: 5-conclusion.tex
In this work, we leveraged rare word definitions in English dictionary to improve language model pre-training. 
When encountering a rare word in the input text during pre-training, we fetched its definition from Wiktionary and appended it to the end of the input text. In order to make better interactions between the input text and rare word definitions, we proposed two novel self-supervised training tasks to help language model learn better representations for rare words.
Experimental on the GLUE benchmark and eight specialized domain datasets demonstrated that our method significantly improved the understanding of rare words and boosted model performance on various downstream tasks.

%% file: appendix.tex
\begin{table*}[ht]
\begin{center}
\caption{Hyperparameters for model pre-training and domain-adaptive pre-training (DAPT).}
\vspace{-0.1in}
\begin{tabular}{ccc}
\toprule
\textbf{Hyperparameter} & \multicolumn{2}{c}{{\textbf{Assignments}}}  \\
\midrule
Pre-training setting & BERT pre-training & Domain adaptive pre-training  \\
\midrule
number of steps & 300K & 12.5K \\
batch size & 2,000 & 2,000 \\
maximum learning rate & 7e-4 & 1e-4 \\
learning rate optimizer & Adam & Adam \\
Adam epsilon & 1e-6 & 1e-6 \\
Adam beta weights & 0.9, 0.98 & 0.9, 0.98 \\
Weight decay & 0.01 & 0.01 \\
Warmup proportion & 0.06 & 0.06 \\
learning rate decay & linear & linear \\
\bottomrule
\end{tabular}
\label{tab:hyper-pretrain}
\end{center}
\end{table*}

% \begin{table*}[ht]
% \begin{center}
% \vspace{-0.1in}
% \begin{tabular}{ccc}
% \toprule
% \textbf{Computing Resources} & \multicolumn{2}{c}{{\textbf{Information}}}  \\
% \midrule
% Pre-training setting & BERT pre-training & Domain adaptive pre-training  \\
% \midrule
% Infrastructure & 32*V100 & 32*V100 \\
% Memory cost (max) & 32GB & 32GB \\
% Time cost (avg) & 72h-80h & 6h-8h \\
% \bottomrule
% \end{tabular}
% \end{center}
% \end{table*}

\subsection{Preliminary: BERT Pre-training}
We use the BERT~\citep{devlin2019bert} model as an example to introduce the basics of the model architecture and training objective of PLMs. 
BERT is developed on a multi-layer bidirectional Transformer~\citep{vaswani2017attention} encoder.
The Transformer encoder is a stack of multiple identical layers, where each layer has two sub-layers: a self-attention sub-layer and a position-wise feed-forward sub-layer. 
The self-attention sub-layer produces outputs by calculating the scaled dot products of queries and keys as the coefficients of the values,
\begin{equation}
    \text {Attention}(Q, K, V) = \text{Softmax}(\frac{QK^T}{\sqrt{d}})V.
\end{equation}
$Q$(Query), $K$(Key), $V$(Value) are the hidden representations produced by the previous self-attention layer and $d$ is the dimension of the hiddens. Transformer also extends the aforementioned self-attention layer to a multi-head self-attention layer version in order to jointly attend to information from different representation subspaces. 
% The multi-head self-attention sub-layer works as follows,
% \begin{equation}
%     \text{Multi-head}(Q, K, V)=\text{Concat}(\text{head}_1, \cdots , \text{head}_H)W^O     
% \end{equation}
% \begin{equation}
%     \text{head}_k = \text{Attention}(QW^Q_k, KW^K_k, VW^V_k),    
% \end{equation}

% where $W^Q_k \in \mathbb{R}^{d\times d_K}$ , $W^K_k \in \mathbb{R}^{d\times d_K}$ , $W^V_k \in \mathbb{R}^{d\times d_V}$ are projection matrices. $H$ is the number of heads. $d_K$ and $d_V$ are the dimensions of the key and value separately.

% Following the self-attention sub-layer, there is a position-wise feed-forward (FFN) sub-layer, which is a fully connected network applied to every position identically and separately. The FFN sub-layer is usually a two-layer feed-forward network with a ReLU activation function in between. Given vectors $\{h_1, \cdots , h_n\}$, a position-wise FFN sub-layer transforms each $h_i$ as FFN$(h_i) = \sigma (h_iW_1+b_1)W_2+b_2$,
% where $W_1$, $W_2$, $b_1$ and $b_2$ are parameters.

BERT uses the Transformer model as its backbone neural network architecture and trains the model parameters with the masked language modeling (MLM) objective on large text corpora. 
In the masked language modeling task, a random sample of the words in the input text sequence is selected.
The selected positions will be either replaced by special token [MASK], replaced by randomly picked tokens or remain the same.
The objective of masked language modeling is to predict words at the masked positions correctly given the masked sentences. 
RoBERTa (robustly optimized BERT approach) is a retraining of BERT with improved training methodologies, 1000\% more data (i.e., 160 GB) and computation power (i.e., 1024 V100 GPUs).
To improve the training procedure, RoBERTa introduces dynamic masking so that the masked token changes during the training epochs. 
Larger batch-training sizes were also found to be more useful in the training procedure.

\begin{table*}[ht]
\begin{center}
\caption{Hyperparameters for model fine-tuning on GLUE and specialized domain benchmarks.}
\vspace{-0.1in}
\begin{tabular}{ccc}
\toprule
\textbf{Hyperparameter} & \multicolumn{2}{c}{{\textbf{Assignments}}}  \\
\midrule
Fine-tuning setting & GLUE benchmark & Specialized domain  \\
\midrule
number of epochs & 5 or 10 & 10 \\
batch size & 24 or 168 & 168 \\
learning rate & 2e-5 & 2e-5 or 3e-5 \\
learning rate optimizer & Adam & Adam \\
Adam epsilon & 1e-6 & 1e-6 \\
Adam beta weights & 0.9, 0.98 & 0.9, 0.98 \\
Dropout & 0.1 & 0.1 \\
Weight decay & 0.01 & 0.01 \\
learning rate decay & linear & linear \\
\bottomrule
\end{tabular}
\label{tab:hyper-finetune}
\end{center}
\end{table*}

\subsection{BERT Pre-training Details}
\label{sec:pre-training-details}

We conducted experiments on pre-training BERT-base with 110M parameters~\citep{devlin2019bert}. BERT-base consists of 12 Transformer layers. For each layer, the hidden size is set to 768 and the number of attention head is set to 12. All models (including BERT-base and Dict-BERT-base) are pre-trained for 300k steps with batch size 2,000 and maximum sequence length 512. We use Adam~\citep{kingma2015adam} as the optimizer, and set its hyperparameter $\epsilon$ to $1e$-$6$ and $(\beta_1, \beta_2)$ to $(0.9, 0.98)$. The peak learning rate is set to $7e$-$4$ with a 10k-step warm-up stage.
We set the dropout probability to $0.1$ and weight decay to $0.01$. All configurations are reported in Table \ref{tab:hyper-pretrain}.

\subsection{Domain Adaptive Pre-training Details}

We conducted experiments on domain adaptive pre-training (DAPT) of BERT-base and RoBERTa-base. 
RoBERTa
% (short for ``Robustly optimized BERT approach'')
~\citep{liu2019roberta} is a retraining of BERT with improved training methodologies, 1000\% more data (i.e., 160 GB) and computation power (i.e., 1024 V100 GPUs).
To improve the training procedure, RoBERTa removes the next sentence prediction task from BERT’s pre-training and introduces dynamic masking so that the masked token changes during the training epochs. 
To train the models, we followed \cite{gururangan2020don} and domain adaptive pre-training for 12.5k steps with batch size 2,000. All other configurations are reported in Table \ref{tab:hyper-pretrain}.

% \begin{table*}[t]
% \begin{center}
% \caption{SOTA performance on domain specialized datasets.}
% \vspace{-0.1in}
% \begin{tabular}{l|ccc}
% \toprule
% % \textbf{Hyperparameter} & \multicolumn{2}{c}{{\textbf{Assignments}}}  \\
% % \midrule
% Datasets & SOTA method & SOTA & Dict-RoBERTa\\
% \midrule
% ChemProt & \cite{lee2020biobert} & 84.6 & 84.5 \\
% RCT & \cite{lee2020biobert} & 87.6 & 87.5\\
% ACL-ARC & \cite{beltagy2019scibert} & 71.0 & 75.3 \\
% SciERC & \cite{beltagy2019scibert} & 81.8 & 82.5\\
% HP & \cite{beltagy2020longformer} & 94.8 & 94.7 \\
% AGNews & \cite{yang2019xlnet} & 95.5 & 94.8 \\
% Helpful & \cite{gururangan2020don} & 68.7 & 70.8 \\
% IMDB & \cite{thongtan2019sentiment} & 96.2 & 95.5 \\
% \bottomrule
% \end{tabular}
% \label{tab:domain-specialized}
% \end{center}
% \end{table*}

\subsection{Fine-tuning Details}
\label{sec:fine-tuning-details}

Following previous work, we search the learning rates during the fine-tuning for each downstream task. The details are listed in Table \ref{tab:hyper-finetune}.
Each configuration is run five times with different random seeds, and
the average of these five results on the validation set is calculated as the final performance of one configuration. We report the best number over all configurations for each task.

\subsection{Evaluation Metrics}
For GLUE, we followed RoBERTa~\citep{liu2019roberta} and reported Matthews correlation for CoLA, Pearson correlation for STS-B, and Accuracy for other tasks. For specialized tasks, we followed \cite{gururangan2020don} and reported Micro-F1 for Chemprot and RCT-20k, and Macro-F1 for other tasks. For WNLaMPro, we followed \cite{schick2020rare} and reported MRR and P@K.

\subsection{Usage of Wiktionary}
\label{sec:wiktionary}

\paragraph{Polysemy in Wiktionary.}

There are plenty of English words having multiple meanings (aka. polysemy). 
If multiple meanings of a word are appended to the input text sequence simultaneously, it may bring noisy information and disrupt the training of the entire language model.

In this work, we do not pay particular attention to the polysemy issue, because for most rare words, they often only have one meaning in the dictionary.
For example, as mentioned in Section \ref{sec:rare_word}, there are a total of 252,581 rare words in the BERT pre-training corpus (i.e., Bookcorpus and Wikipedia). Among them, 228,658 (90.52\%) words only have one meaning, and 21,721 (8.6\%) words have two meanings. So, words less than three meanings account for more than 99\% of words. To deal with the words having more than one meaning, we use the definition of their first etymology, i.e., the most commonly used meaning, in the Wiktionary.

\paragraph{Rare words during fine-tuning.} 

One important advantage of Dict-BERT is that it can dynamically adjust the vocabulary of rare words, obtain and represent their definitions in a dictionary in a plug-and-play manner.
As different domain datasets usually follow different word distributions, the pre-trained language model may still encounter many rare words when fine-tuned on the downstream tasks.
To enhance the rare word representations during fine-tuning process, when encountering a rare word in the input text sequence, we append its definition to the end of input text sequence, just like what we did in pre-training.

\paragraph{Negative words sampling.} 

The sentence-level definition discrimination task needs to sample negative words for each input text sequence.
In order to select harder negative words, we explored several different strategies such as using words with similar GloVe~\cite{pennington2014glove} embeddings or taking the synonyms of rare words provided in the dictionary as negative samples.
However, either of the two methods has the problem of extremely low coverage. Most of the rare words are not appeared in GloVe, and only about 10\% of the words have synonyms in the Wiktionary.
Thus, we chose to use the random negative sampling strategy.

% \subsection{SOTA on Domain Specialized Datasets}

% Table \ref{tab:domain-specialized} reported other state-of-the-art performance on eight domain specialized datasets. These SOTA language models are either trained on domain specific corpus (e.g., SciBERT~\cite{beltagy2019scibert}, BioBERT~\cite{lee2020biobert}) or have much larger size than other Dict-BERT/Dict-RoBERTa (e.g., XLNet-large~\cite{yang2019xlnet}, LongFormer~\cite{beltagy2020longformer}).

% Compared with these SOTA models, Dict-BERT can still outperform SciBERT on two computer science datasets, while achieving on par performance on other six domain specialized datasets.